\documentclass[letterpaper,10pt,conference,twoside]{IEEEtran}
\IEEEoverridecommandlockouts

\usepackage[inline]{enumitem}
\usepackage{amsmath}
\usepackage{amsfonts}
\usepackage{amssymb}
\usepackage{algorithmic}
\usepackage{algorithm}
\usepackage{array}
\usepackage{textcomp}
\usepackage{stfloats}
\usepackage{verbatim}
\usepackage{graphicx}
\usepackage[font=footnotesize]{caption}
\usepackage[font=footnotesize]{subcaption}
\usepackage[noadjust]{cite}

\usepackage{url}

\makeatletter
\let\NAT@parse\undefined
\makeatother
\usepackage[pagebackref=true,breaklinks=true,colorlinks,bookmarks=false]{hyperref}
\makeatletter
\newcommand*{\textlabel}[2]{%
  \edef\@currentlabel{#1}
  \phantomsection
  #1\label{#2}
}
\makeatother

\usepackage{cleveref}[2012/02/15]
\crefformat{footnote}{#2\footnotemark[#1]#3}
\usepackage{textpos}
\usepackage{amsthm}
\usepackage{xcolor}
\usepackage{tikz}
\usepackage[scaled]{helvet}
\usepackage{flushend}

\theoremstyle{definition}

\newtheorem{defn}{Definition}[section]

\newtheorem{pb}{Problem}[section]

\makeatletter
\newcommand\notsotiny{\@setfontsize\notsotiny\@vipt\@viipt}
\makeatother

\begin{document}
\bstctlcite{IEEEexample:BSTcontrol}

\title{\LARGE\bf Energy-Aware Ergodic Search: Continuous Exploration for Multi-Agent Systems with Battery Constraints}

\author{Adam Seewald, Cameron J. Lerch, Marvin Chanc{\'a}n, Aaron M. Dollar, and Ian Abraham
  \thanks{This work was partly supported by Yale University and a gift from the Boston Dynamics AI Institute.}
  \thanks{A.\hspace*{.4ex}S., C.\hspace*{.4ex}J.\hspace*{.4ex}L., M.\hspace*{.4ex}C., A.\hspace*{.4ex}M.\hspace*{.4ex}D., and I.\hspace*{.4ex}A. are with the Department of Mechanical Engineering and Materials Science, Yale University, CT, USA. Email: {\tt\footnotesize \href{mailto:adam.seewald@yale.edu}{adam.seewald@yale.edu};}}
}

\maketitle

\vspace*{-.5cm}
\begin{abstract} 
  Continuous exploration without interruption is~important in scenarios such as search and rescue and precision agriculture, where consistent presence is needed to detect events over large areas. Ergodic search already derives continuous trajectories in these scenarios so that a robot spends more time in areas with high information density. However, existing literature on ergodic search does not consider the robot's energy constraints, limiting how long a robot can explore. In fact, if the robots are battery-powered, it is physically not possible to continuously explore on a single battery charge. Our paper tackles this challenge, integrating ergodic search methods with energy-aware coverage. We trade off battery usage and coverage quality, maintaining uninterrupted exploration by at least one agent. Our approach derives an abstract battery model for future state-of-charge estimation and extends canonical ergodic search to ergodic search under battery constraints. Empirical data from simulations and real-world experiments demonstrate the effectiveness of our energy-aware ergodic search, which ensures continuous 
  exploration and guarantees spatial coverage.
\end{abstract}

\vspace*{-.3cm}
\section{Introduction}
\noindent
Robotic exploration is a recurring problem in different scenarios, e.g., inspection, surveying, etc. It 
typically involves coverage path planning (CPP), which deals with deriving robots' trajectories that traverse every point in a given space~\cite{choset2001coverage,galceran2013survey,cabreira2019survey}. Within CPP, ergodic search is a recent and promising direction~\cite{abraham2021ergodic,miller2016ergodic,dressel2018optimality,torre2016ergodic,shetty2022ergodic,prabhakar2020ergodic,coffin2022multi,lerch2023safety,abraham2018decentralized,patel2021multi,dong2023time,abraham2017ergodic,rao2023multi,ayvali2017ergodic}, as it enhances the efficiency of traditional CPP by optimizing the time a robot spends in a given region w.r.t. an information measure. As a result, ergodic search derives trajectories so that the robots spend more time in areas with high information density while quickly passing areas with low information density~\cite{mathew2011metrics,
patel2021multi}. The user can specify areas of interest, e.g., where the robots should spend more time exploring in a search and rescue scenario~\cite{dong2023time}, where the robots should collect more data in a precision agriculture scenario~\cite{rao2023multi},~etc.

Canonical ergodic search already derives continuous exploration trajectories~\cite{
miller2013trajectory,miller2016ergodic,abraham2017ergodic}, but it is physically not possible for robots to continue exploring on a single battery charge. 
Scenarios involving CPP, however, often require that the space is covered continuously.
This work enhances the current ergodic search literature by incorporating more traditional energy-aware CPP approaches~\cite{difranco2015energy,difranco2016coverage,shnaps2016online,cabreira2018energy,wei2018coverage,jensen2021near}, battery- and energy-aware planning~\cite{mei2004energy,mei2005case,kim2005energy,seewald2022energy}, and planning of energy trade-offs~\cite{ondruska2015scheduled,sudhakar2020balancing}. It answers the question: \textit{Is it possible to tradeoff battery and coverage quality so that there is at least one agent exploring 
at all times}?

Prior literature has studied ergodic search in manipulation~\cite{shetty2022ergodic}, tactile sensing~\cite{abraham2017ergodic}, stochastic dynamics~\cite{ayvali2017ergodic,torre2016ergodic}, distributed information~\cite{miller2016ergodic}, time optimality~\cite{dong2023time}, and active learning~\cite{abraham2021ergodic}. Ergodic search for multi-agent systems~\cite{prabhakar2020ergodic,coffin2022multi} has been applied in conjunction with low-information sensors~\cite{coffin2022multi,lerch2023safety,abraham2018decentralized}, swarms control~\cite{prabhakar2020ergodic}, obstacle avoidance~\cite{lerch2023safety}, and decentralized systems~\cite{abraham2018decentralized}. Ergodic search has been proven successful in use cases involving urban environments~\cite{patel2021multi} and information gathering~\cite{dressel2018optimality}.
While prior literature includes ergodic search methods in a variety of settings, energy constraints have not been studied yet. 
Partly due to these constraints, the uninterrupted exploration 
that 
considers a spatial distribution is currently hindered. 

\begin{figure}[t!]
  \centering
  \vspace*{-.1cm}
  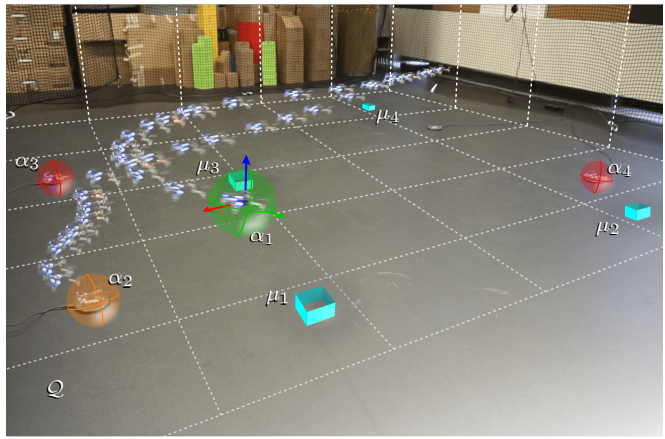
  \caption{\textbf{Example of energy-aware ergodic search}. A set of agents explores $\mathcal{Q}$, focusing on areas with high information density $\mu_1$, $\mu_2$, $\mu_3$, and $\mu_4$, employing ergodic search. The exploration is continuous and uninterrupted 
  so that there is always one agent exploring -- $\alpha_1$, whereas $\alpha_2$, $\alpha_3$, and $\alpha_4$ are recharging. The colors of the spheres indicate the state of charge.}
  \label{fig:abs}
  \vspace*{-.4cm}
\end{figure}

Our approach derives an abstract battery model~\cite{zhao2017observability} for battery state of charge (SoC) estimation at future time instances. We first adapt canonical ergodic search to multi-agent ergodic search~\cite{prabhakar2020ergodic,coffin2022multi}. We then utilize the formulation to propose energy-aware ergodic search, i.e., ergodic search under battery constraints. The exploration is continuous and uninterrupted, employing a finite horizon framework reminiscent of a model predictive controller~\cite{seewald2022energy}. 
Experimental data from simulations and real-world experiments show 
our newly proposed energy-aware ergodic search. 
We show with empirical evidence that we can effectively explore a space, 
there is 
at least one agent exploring, the spatial distribution is satisfied, and the exploration is 
uninterrupted in Section~\ref{sec:res}. 
Figure~\ref{fig:abs} shows our experimental setup, which resembles a search and rescue scenario. Four agents explore the space where the high information density is represented by cyan boxes. Some agents are actively exploring (green sphere), while others are grounded and recharging (red and orange spheres). 
The detailed results from the experimental evaluation and the code to replicate our approach are made available on the project repository webpage\footnote{\label{reflink}{\tt\footnotesize\href{https://github.com/adamseew/enerergo}{github.com/adamseew/enerergo}}}.

The remainder of the paper is then structured as follows. Sec.~\ref{sec:pb} formulates the problem of energy-aware ergodic search. Sec.~\ref{sec:meth} discusses the methods for both the canonical ergodic search and our battery model enhanced ergodic search. Sec.~\ref{sec:conc} concludes and proposes future directions.

\section{Problem Formulation}\label{sec:pb}
\noindent
This work addresses the problem of exploring a bounded space with multiple agents, proportionally to a spatial distribution and in such a way that there is at least one agent exploring. 
In the remainder of the text, we will use the terms \textit{continuously} and \textit{uninterruptedly} to indicate that there is at least one agent exploring the space at all times, i.e., a certain measurable level of coverage is always satisfied. 
Canonical ergodic search~\cite{mathew2011metrics,miller2013trajectory,abraham2021ergodic,miller2016ergodic,dressel2018optimality,torre2016ergodic,shetty2022ergodic} does not deal with uninterrupted exploration. It derives an agent's control -- or analogously multiple agents' control~\cite{prabhakar2020ergodic,coffin2022multi,lerch2023safety,abraham2018decentralized,patel2021multi} -- so that its trajectory maximizes an ergodic metric 
in the spectral~domain. 
Let us thus define the concepts of ergodicity and ergodic metric.

\begin{defn}[Ergodicity]
  Consider a bounded space $\mathcal{Q}\subset\mathbb{R}^D$ of dimension $D\in\mathbb{N}_{>0}$. 
  A trajectory $\mathbf{q}(t)\in\mathcal{Q}$ is \textit{ergodic} with respect to a spatial distribution $\phi$, i.e., is distributed among regions of high expected distribution~\cite{miller2016ergodic}, if and only~if
  \begin{equation}
    \lim_{t\rightarrow \infty}{\small
      \frac{1}{t}\int_{\mathcal{T}}\Omega\bigg(\kappa\big(\mathbf{q}(t)\big)\bigg)=\int_{\mathcal{Q}}\phi(\mathbf{q})\,\Omega(\mathbf{q})\,d\mathbf{q},
    }
  \end{equation}
  for all Lebesgue functions $\Omega$~\cite{mathew2011metrics}. Here, the function $\kappa$ maps the state to the exploration workspace.
  \end{defn}

\begin{defn}[Ergodic metric]\label{def:ergom}
  Consider a time average distribution that describes where the robot spends more time over a finite time window $[t_0,t_f]$, i.e., $h\big(\mathbf{q}(t)\big)=\int_{\mathcal{T}}{\Delta\big(\mathbf{q}(t)\big)}\,dt\,/(t_f-t_0)$, $\Delta$ is a Dirac delta function, and $t_0,t_f\in\mathbb{R}_{>0}$ are the initial and final time instants. 
  An \textit{ergodic metric} is defined as the $L^2$ inner product between the time average distribution $h$ and the average of the spatial distribution $\phi$.
\end{defn}

\begin{pb}[Ergodic search]\label{pb:ergo}
  Consider the bounded space $\mathcal{Q}$ and a spatial distribution $\phi$ s.t. $\int_{\mathcal{Q}}\phi\,d\mathbf{q}=1,\,\phi(\mathbf{q})\geq 0,\,\, \forall \mathbf{q}\in\mathcal{Q}$.
  \textit{Ergodic search problem} is the problem of~deriving a control action $\mathbf{u}(t)\in\mathcal{U}\subset\mathbb{R}^V$ with $V\in\mathbb{N}_{>0}$ so that the ergodic metric is minimized (see Definition~\ref{def:ergom}). 
\end{pb}

We derive an ergodic metric in Equation~(\ref{eq:ergmetric}) in Sec.~\ref{sec:ergosearch}. 
The notation $\mathbb{R}$ and $\mathbb{N}$ indicates reals and naturals, $\mathbb{N}_{>0}$ strictly naturals. Bold notation is used for vectors.

Let us extend the canonical ergodic search problem to energy-aware ergodic search, i.e., uninterrupted multi-agent exploration 
under spatial distribution and battery~constraints.

\begin{pb}[Energy-aware ergodic search]\label{pb:enerergo}
  Consider a set of $n$ agents $\boldsymbol{\alpha}:=\{\alpha_1,\alpha_2,\dots,\alpha_n\}$, a bounded space $\mathcal{Q}$, and a spatial distribution $\phi$ similar to Problem~\ref{pb:ergo}.~\textit{Energy-aware ergodic search problem} is the problem of deriving each agent's ${}^j\alpha$ control action ${}^j\mathbf{u}(t)$ so that $\forall j,\,\,{}^j\mathbf{q}(t)$ induces a time average distribution $h\big(\mathbf{q}(t)\big)$ that is proportional to the spatial distribution $\phi$ and 
  the ergodic metric (calculated between the agents' averaged time average distribution in Definition~\ref{def:ergom}) is satisfied, i.e., there is an upper bound on the ergodic metric 
  for a given $\gamma\in\mathbb{R}_{>0}$.
\end{pb}

We will provide a solution to Problem~\ref{pb:enerergo} (see Sec.~\ref{sec:meth}), assuming that there are one or more areas in $\mathcal{Q}$ -- namely, charging stations -- where the agents can land and recharge the battery, using, e.g., wireless charging (see Sec.~\ref{sec:res}).

\section{Methods}\label{sec:meth}
\noindent
In this section, we discuss the methods utilized in this work for continuous, uninterrupted exploration with multiple agents and proportionally to a spatial distribution. 
We discuss how to achieve the latter in Sec.~\ref{sec:ergosearch} and the former in Sec.~\ref{sec:batt}.

\subsection{Ergodic search}\label{sec:ergosearch}
\noindent
To derive an agent's trajectory proportionally to a spatial distribution, canonical ergodic search first requires defining the distribution $\phi$. 
For this purpose, in both Problem~\ref{pb:ergo} and \ref{pb:enerergo}, let us consider a Gaussian mixture model (GMM)
\begin{equation}\label{eq:gmm}
  \phi(\boldsymbol{\delta},\mathbf{q}):=\sum_{k=1}^{m}\delta_k\,\mathcal{N}(\mathbf{q}\,|\,\mu_k,\Sigma_k),
\end{equation} 
composed of $m$ Gaussians $\mathcal{N}$. Each has a covariance matrix ${\Sigma_k}$ $\in\mathbb{R}^{D\times D}$, center $\mu_k\in\mathcal{Q}$, and positive mixing coefficient $\delta_k\in\boldsymbol{\delta}$ such that, for each $k$, the sum of $\delta_k$ is $\leq 1$. 

The goal of ergodic search is to minimize an ergodic metric (see Definition~\ref{def:ergom}) such as~\cite{mathew2011metrics}
\begin{equation}\label{eq:ergmetric}
  \mathcal{E}(\boldsymbol{\delta},\mathbf{q}(t)):=\frac{1}{2}\sum_{k\in\mathcal{K}}\Lambda_k \big( c_k(\mathbf{q}(t))-\phi_k(\boldsymbol{\delta}) \big)^2,
\end{equation}
where $\phi_k$ are coefficients derived utilizing the Fourier series on the spatial distribution $\phi$ and $c_k$ on the trajectory $\mathbf{q}(t)$. They are detailed in Eq.~(\ref{eq:phik}) and (\ref{eq:ck}) respectively.
An agent whose trajectory 
minimizes Eq.~(\ref{eq:ergmetric}) for $t\rightarrow\infty$ is \textit{optimally ergodic} w.r.t. $\phi$~\cite{abraham2018decentralized}. 

$\Lambda_k$ is a weight factor. That is, if 
\begin{equation}
  \Lambda_k=(1+\lVert k\rVert^2)^{(-D-1)/2},
\end{equation}
lower frequencies have more weight~\cite{miller2016ergodic}.
$\mathcal{K}\in\mathbb{N}^D$ is a set of index vectors that covers $[K]\times\cdots\times[K]\in\mathbb{N}^{K^D}$ 
where $K$ is a given number of frequencies with the fundamental frequency~\cite{calinon2020mixture}. The notation $[K]$ indicates positive naturals up to $K$.

The coefficients $c_k$ are derived using the Fourier series basis function. If we consider the trigonometric form, they can be expressed
\begin{equation}\label{eq:ck}
  \begin{split}
    c_k(\mathbf{q}(t)):=\int_{\mathcal{T}}\frac{1}{L^D}\prod_{d\in[D]_{>0}\hspace*{-1ex}}\big(& \cos(k_d\,\mathbf{q}_d(\tau)\,\psi)\\[-2ex]
    &\vspace*{-4ex}-i\sin(k_d\,\mathbf{q}_d(\tau)\,\psi) \big)\,d\tau/t,
  \end{split}
\end{equation}
where $\psi$ is $2\pi/L$ for a given period $L\in\mathbb{R}_{>0}$, $i$ is the imaginary unit, $k_d$ is the $d$th item of $k$, and $\mathbf{q}_d$ is the $d$th item of $\mathbf{q}$.

The interval $\mathcal{T}$ is built so that the integration is between $\tau=t_0$ and $t$, and the notation $[D]_{>0}$ indicates strictly positive naturals up to $D$.


To derive the coefficients $\phi_k$, let us consider the GMM model in Eq.~(\ref{eq:gmm}) on a search space $\mathcal{Q}$. The space is further bounded to a symmetric set $[-L/2,L/2]^D$ since the Gaussians are symmetric about the zero axes. 
As a result, the model can be expressed~\cite{calinon2020mixture}
\begin{equation}
  \Phi(\boldsymbol{\delta},\mathbf{q}):=\sum_{d\in[2^D]_{>0}}\sum_{k=1}^{m}\delta_k\,\mathcal{N}(\mathbf{q}\,|\,A_d\mu_k,A_d\Sigma_k A_d^T)/2^D,
\end{equation}
where $A_d\in\mathbb{R}^{D\times D}$ are linear transformation matrices. 

Let us call the integrand in Eq.~(\ref{eq:ck}) $c$
. It maps the space to the spectral domain. The equivalent of Eq.~(\ref{eq:ck}) for the spatial distribution can be then expressed
\begin{equation}\label{eq:phik}
  \phi_k(\boldsymbol{\delta}):=\int_{\mathcal{Q}} \Phi(\boldsymbol{\delta},\mathbf{q})\,c(\mathbf{q})\,\,d\mathbf{q}.
\end{equation}

The space $\mathcal{Q}$ is built so that the integration is within the points of the bounded symmetric set $\mathbf{q}\in[-L/2,L/2]^D$.

Both the coefficients $c_k$ and $\phi_k$ are evaluated per each $k$ in $\mathcal{K}$ in Eq.~(\ref{eq:ergmetric}).

Let us first formulate the solution to Problem~\ref{pb:ergo}, utilizing a formulation borrowed from canonical ergodic search.
If the agent's dynamics is described by a generic differential equation $\dot{\mathbf{q}}(t)=f\big({\small\mathbf{q}(t),}$ ${\small\mathbf{u}(t)})$, an optimal control problem (OCP) that selects an ergodic control can be formulated~\cite{ayvali2017ergodic}
\begin{subequations}\vspace*{-.4cm}\label{eq:ocpergo}\begin{align}
  \min_{\mathbf{q}(t),\mathbf{u}(t)}&\int_{\mathcal{T}}\mathbf{u}(\tau)^TR\mathbf{u}(\tau)\,d\tau+{\mathcal{E}(\boldsymbol{\delta},\mathbf{q}(t))},\label{eq:ocpergomin}\\
  \text{s.t. }\dot{\mathbf{q}}&=f(\mathbf{q}(t),\mathbf{u}(t)),\\
  \mathbf{q}&(t)\in\mathcal{Q},\,\mathbf{u}(t)\in\mathcal{U},\\
  \mathbf{q}&(t_0), \mathbf{q}(t_f)\text{ are given},\label{eq:ocpconsttotf}
\end{align}\end{subequations}
where the ergodic metric is derived in Eq.~(\ref{eq:ergmetric}), $R\in\mathbb{R}^{V\times V}$ is a control penalizing diagonal positive-definite matrix, and $t_0, t_f$ are the first and last time instants respectively. 
The interval $\mathcal{T}$ is $[t_0, t_f]$.

To formulate the solution to Problem~\ref{pb:enerergo}, let us first extend the OCP in Eq.~(\ref{eq:ocpergo}) to multi-agent systems. Eq.~(\ref{eq:ocpergomin}) becomes
\begin{equation}\label{eq:ocpergomulti}
  \min_{\boldsymbol{\Theta}}\,\,\,{\frac{1}{n}\left(\sum_{k=1}^{n}\int_{\mathcal{T}_k}{}^k\hspace*{-.1ex}\mathbf{u}(\tau)^TR_k\,{}^k\hspace*{-.1ex}\mathbf{u}(\tau)\,d\tau\right)+\mathcal{E}(\boldsymbol{\delta},\boldsymbol{\Theta}_{\mathbf{q}})},
\end{equation}
where 
the control penalizing term $R_k$ 
is now agent-specific. The term $\boldsymbol{\Theta}_{\mathbf{q}}$ is ${}^1\mathbf{q}(t),{}^2\mathbf{q}(t),$ $\dots,{}^n\mathbf{q}(t)$, and $\boldsymbol{\Theta}$ is $\boldsymbol{\Theta}_{\mathbf{q}},{}^1\mathbf{u}(t),{}^2\mathbf{u}(t),\dots{}^n\mathbf{u}(t)$. $\mathcal{T}_k$ is $[{}^kt_0, {}^kt_f]$, i.e., different agents might have different durations.

The ergodic metric is now evaluated for multiple agents\vspace*{-.1cm}
\begin{equation}\label{eq:multiergometric}
  \mathcal{E}(\boldsymbol{\delta},\boldsymbol{\Theta}_{\mathbf{q}}):={\small
  \frac{1}{2}
  \sum_{k\in\mathcal{K}}\Lambda_k 
    \Bigg(
      \frac{1}{n}\sum_{j\in[n]} c_k\big({}^j\mathbf{q}(t)\big)-\phi_k(\boldsymbol{\delta})
    \Bigg)^2\hspace*{-.5ex}.}
\end{equation}
\vspace*{-.2cm}

Let us consider a vector $\mathbf{b}\in\mathbb{R}^3$, detailed later in Sec.~\ref{sec:batt}, whose trajectory $\mathbf{b}(t)$ describes the evolution of some battery metrics' in time. If $\mathbf{b}_{\text{SoC}}$ is the value of the vector that expresses the battery 
SoC, the expression in Eq.~(\ref{eq:ocpergomulti}) might select ergodic metrics corresponding to trajectories that are impossible to traverse in the $\mathbf{b}_{\text{SoC}}\in(0,1]$ domain. 
In order to satisfy the battery SoC domain and always keep at least one agent exploring, an OCP must satisfy an additional constraint
\begin{equation}\label{eq:ocpbattconst}
  \exists k\in[n]\text{ s.t. }{}^k\mathbf{b}_{\text{SoC}}(t_f)\in(0,b_f],
\end{equation}
where $b_f\in(0,1]\subset\mathbb{R}_{>0}$ is a given desired battery SoC at the final time instant.

Finally, let us consider the realistic assumption that the optimization horizon 
is known and is,
e.g., an empirically collected value that corresponds to one of the agents' discharge times (see Sec.~\ref{sec:res}).

The OCP that provides a solution to Problem~\ref{pb:enerergo} can be formulated 
\begin{subequations}\label{eq:ocpfinal}\begin{align}
  \min_{\boldsymbol{\Theta}}\,\,\,&{\frac{1}{n}\sum_{k=1}^{n}\int_{\mathcal{T}_k}{}^k\hspace*{-.1ex}\mathbf{u}(\tau)^TR_k\,{}^k\hspace*{-.1ex}\mathbf{u}(\tau)\,d\tau},\label{eq:ocpfinalcost}\\
  \text{\hspace*{.2ex}s.t.\hspace*{.3ex}}{}^1\hspace*{-.4ex}\dot{\mathbf{q}}&(t)\hspace*{-.4ex}=\hspace*{-.6ex}f_1({}^1\hspace*{-.2ex}\mathbf{q}(t),\hspace*{-.8ex}{}^1\hspace*{-.2ex}\mathbf{u}(t)),\hspace*{-.3ex}{\small\dots},\hspace*{-.7ex}{}^n\hspace*{-.3ex}\dot{\mathbf{q}}(t)\hspace*{-.3ex}=\hspace*{-.5ex}f_n({}^n\hspace*{-.4ex}\mathbf{q}(t),\hspace*{-.7ex}{}^n\hspace*{-.4ex}\mathbf{u}(t)),\\
  {}^1\hspace*{-.2ex}\mathbf{q}&(t),\dots,{}^n\hspace*{-.2ex}\mathbf{q}(t)\in\mathcal{Q},\,{}^1\hspace*{-.2ex}\mathbf{u}(t),\dots,{}^n\hspace*{-.2ex}\mathbf{u}(t)\in\mathcal{U},\\
  \exists &k\in[n]\text{ s.t. }{}^k\mathbf{b}_{\text{SoC}}(t_f)\in(0,b_f],\label{eq:battconst}\\
  \mathcal{E}&(\boldsymbol{\delta},\hspace*{-.2ex}\boldsymbol{\Theta}_{\mathbf{q}})\leq\gamma,\label{eq:ergoconst}\\
  g&{}_1(\boldsymbol{\delta},\hspace*{-.6ex}{}^1\hspace*{-.3ex}\mathbf{q}(t),\hspace*{-.6ex}{}^1\hspace*{-.3ex}\mathbf{u}(t))\leq 0,\dots,g_n(\boldsymbol{\delta},\hspace*{-.6ex}{}^n\hspace*{-.3ex}\mathbf{q}(t),\hspace*{-.6ex}{}^n\hspace*{-.3ex}\mathbf{u}(t))\hspace*{-.3ex}\leq\hspace*{-.3ex}0,\label{eq:ocpfinalextra1}\\
  {}^1\hspace*{-.4ex}\mathbf{q}&(t_0),{}^1\hspace*{-.4ex}\mathbf{q}(t_f),\dots,{}^n\hspace*{-.4ex}\mathbf{q}(t_0),{}^n\hspace*{-.4ex}\mathbf{q}(t_f),b_f,\gamma\text{ are given},\label{eq:ocpgivenconst}
\end{align}\end{subequations}
where constraints in Eq.~(\ref{eq:ocpfinalextra1}) are optional and express additional requirements 
(see Sec.~\ref{sec:res}). 
The constraint in Eq.~(\ref{eq:battconst}) ensures that there is at least one agent exploring at all time instants in the optimization horizon. 
The ergodic metric is integrated into the constraint as proposed in~\cite{dong2023time}, and 
the evolutions of the agents' states in time are described by generic differential equations ${}^k\dot{\mathbf{q}}(t)=f_k($ ${}^k\mathbf{q}(t),{}^k\mathbf{u}(t))$.


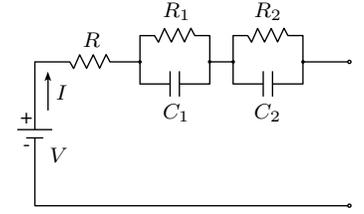
\begin{figure}[t!]
  \vspace*{-.2cm}
  \begin{minipage}[c]{.43\columnwidth}
    \vspace*{.24cm}
    \caption{\textbf{Abstract equivalent circuit model} for state-of-charge estimation~\cite{seewaldphdthesis}. The model consists of a second-order resistor-capacitor circuit with two resistors $R_1$ and $R_2$ and two capacitors $C_1$ and $C_2$ in two separate circuit elements. An additional resistor $R$ is also employed. 
    } 
    \label{fig:thevenin}
  \end{minipage}
  \begin{minipage}[c]{.57\columnwidth}
    \centering
    \vspace*{-.1cm}
    \def \globalscale {.600000}
\begin{tikzpicture}[y=0.80pt, x=0.80pt, yscale=-\globalscale, xscale=\globalscale, inner sep=0pt, outer sep=0pt]
\footnotesize
\path[draw=black,line join=round,line width=0.512pt] (0.3200,96.5242) -- (27.8896,96.5242);

\path[draw=black,line join=round,line width=0.512pt] (7.3265,103.0430) -- (20.8829,103.0430);

\path[draw=black,line join=round,line width=0.512pt] (14.2088,96.5048) -- (14.2088,42.8600);

\path[draw=black,line join=round,line width=0.512pt] (260.9630,156.5580) -- (14.2088,156.5580) -- (14.2088,102.9130);

\path[draw=black,line join=round,line width=0.512pt] (260.8000,43.0793) -- (225.4320,43.0790);

\path[draw=black,line join=round,line width=0.512pt] (97.0029,43.0790) -- (73.3821,43.0790) -- (70.3453,37.3038) -- (65.2383,48.0961) -- (60.2246,37.4905) -- (54.9619,48.1735) -- (49.5124,37.4750) -- (44.7480,48.3298) -- (41.7355,43.1054) -- (13.9755,43.1054);

\path[draw=black,line join=round,line width=0.512pt] (262.0370,155.2530) .. controls (262.7240,155.2530) and (263.2800,155.8100) .. (263.2800,156.4960) .. controls (263.2800,157.1820) and (262.7240,157.7390) .. (262.0370,157.7390) .. controls (261.3510,157.7390) and (260.7950,157.1820) .. (260.7950,156.4960) .. controls (260.7950,155.8100) and (261.3510,155.2530) .. (262.0370,155.2530) -- cycle;

\path[draw=black,line join=round,line width=0.512pt] (262.0370,41.8587) .. controls (262.7240,41.8587) and (263.2800,42.4151) .. (263.2800,43.1014) .. controls (263.2800,43.7878) and (262.7240,44.3441) .. (262.0370,44.3441) .. controls (261.3510,44.3441) and (260.7950,43.7878) .. (260.7950,43.1014) .. controls (260.7950,42.4151) and (261.3510,41.8587) .. (262.0370,41.8587) -- cycle;

\path[cm={{1.0,0.0,0.0,1.0,(3.0,92.0)}}] (0.0000,0.0000) node[above right] () {+};

\path[cm={{1.0,0.0,0.0,1.0,(5.0,113.0)}}] (0.0000,0.0000) node[above right] () {-};

\path[cm={{1.0,0.0,0.0,1.0,(26.0,122.0)}}] (0.0000,0.0000) node[above right] () {$V$};

\path[cm={{1.0,0.0,0.0,1.0,(52.0,31.0)}}] (0.0000,0.0000) node[above right] () {$R$};

\path[draw=black,line join=round,line width=0.512pt] (24.5779,81.3732) -- (24.5779,55.7215);

\path[fill=black,line join=round,line width=0.160pt] (21.7737,58.1955) -- (24.5247,56.6509) -- (27.1298,58.1885) -- (24.4458,50.7447) -- (21.7737,58.1955) -- cycle;

\path[cm={{1.0,0.0,0.0,1.0,(31.0,73.0)}}] (0.0000,0.0000) node[above right] () {$I$};

\path[draw=black,line join=round,line width=0.512pt] (97.0247,60.6763) -- (97.0247,22.1411);

\path[draw=black,line join=round,line width=0.512pt] (96.7722,60.5081) -- (120.7290,60.5085);

\path[draw=black,line join=round,line width=0.512pt] (128.0790,60.5086) -- (152.0390,60.5091) -- (152.0390,22.3621) -- (140.1990,22.3602) -- (137.1630,16.5849) -- (132.0560,27.3773) -- (127.0420,16.7717) -- (121.7800,27.4547) -- (116.3300,16.7562) -- (111.5660,27.6108) -- (108.5530,22.3864) -- (96.7913,22.3869);

\path[draw=black,line join=round,line width=0.512pt] (120.7490,70.4544) -- (120.7490,50.4562);

\path[draw=black,line join=round,line width=0.512pt] (128.0730,70.4544) -- (128.0730,50.4562);

\path[draw=black,line join=round,line width=0.512pt] (170.4000,43.0569) -- (152.1120,43.0573);

\path[draw=black,line join=round,line width=0.512pt] (170.4220,60.6542) -- (170.4220,22.1190);

\path[draw=black,line join=round,line width=0.512pt] (170.1690,60.4860) -- (194.1260,60.4864);

\path[draw=black,line join=round,line width=0.512pt] (201.4760,60.4865) -- (225.4360,60.4870) -- (225.4360,22.3400) -- (213.5960,22.3381) -- (210.5600,16.5628) -- (205.4530,27.3552) -- (200.4400,16.7496) -- (195.1770,27.4326) -- (189.7270,16.7341) -- (184.9630,27.5887) -- (181.9510,22.3643) -- (170.1890,22.3648);

\path[draw=black,line join=round,line width=0.512pt] (194.1470,70.4323) -- (194.1470,50.4341);

\path[draw=black,line join=round,line width=0.512pt] (201.4700,70.4323) -- (201.4700,50.4341);

\path[draw=black,fill=black,line join=round,line width=0.512pt] (225.4900,42.2159) .. controls (225.9770,42.2159) and (226.3720,42.6107) .. (226.3720,43.0978) .. controls (226.3720,43.5850) and (225.9770,43.9799) .. (225.4900,43.9799) .. controls (225.0030,43.9799) and (224.6080,43.5850) .. (224.6080,43.0978) .. controls (224.6080,42.6107) and (225.0030,42.2159) .. (225.4900,42.2159) -- cycle;

\path[draw=black,fill=black,line join=round,line width=0.512pt] (170.4530,42.1766) .. controls (170.9400,42.1766) and (171.3350,42.5714) .. (171.3350,43.0585) .. controls (171.3350,43.5457) and (170.9400,43.9406) .. (170.4530,43.9406) .. controls (169.9660,43.9406) and (169.5710,43.5457) .. (169.5710,43.0585) .. controls (169.5710,42.5714) and (169.9660,42.1766) .. (170.4530,42.1766) -- cycle;

\path[draw=black,fill=black,line join=round,line width=0.512pt] (151.9710,42.1544) .. controls (152.4580,42.1544) and (152.8530,42.5491) .. (152.8530,43.0363) .. controls (152.8530,43.5235) and (152.4580,43.9184) .. (151.9710,43.9184) .. controls (151.4830,43.9184) and (151.0890,43.5235) .. (151.0890,43.0363) .. controls (151.0890,42.5491) and (151.4830,42.1544) .. (151.9710,42.1544) -- cycle;

\path[draw=black,fill=black,line join=round,line width=0.512pt] (97.0363,42.2150) .. controls (97.5234,42.2150) and (97.9183,42.6099) .. (97.9183,43.0971) .. controls (97.9183,43.5843) and (97.5234,43.9792) .. (97.0363,43.9792) .. controls (96.5491,43.9792) and (96.1542,43.5843) .. (96.1542,43.0971) .. controls (96.1542,42.6099) and (96.5491,42.2150) .. (97.0363,42.2150) -- cycle;

\path[cm={{1.0,0.0,0.0,1.0,(115.0,10.0)}}] (0.0000,0.0000) node[above right] () {$R_1$};

\path[cm={{1.0,0.0,0.0,1.0,(187.0,10.0)}}] (0.0000,0.0000) node[above right] () {$R_2$};

\path[cm={{1.0,0.0,0.0,1.0,(115.0,90.0)}}] (0.0000,0.0000) node[above right] () {$C_1$};

\path[cm={{1.0,0.0,0.0,1.0,(187.0,90.0)}}] (0.0000,0.0000) node[above right] () {$C_2$};

\end{tikzpicture}
  \end{minipage}
  \vspace*{-.4cm}
\end{figure}

\subsection{Battery modeling}\label{sec:batt}
\noindent
To derive a battery model for continuous exploration -- a model that allows us to predict when an agent is exploring and when it conversely should be recharging the battery -- let us consider an abstract equivalent circuit model (ECM). These models are commonly employed in battery metrics estimation for robots and other applications, especially if equipped with rechargeable battery cells~\cite{zhang2018online,xiaosong2012comparative,hasan2018exogenous,hinz2019comparison,mousavi2014various,seewald2022energy}.

The ECM model we employ is a second-order resistor-capacitor (RC) circuit model, as illustrated in Fig.~\ref{fig:thevenin}~\cite{seewaldphdthesis}. 
Formally, it can be expressed~\cite{zhao2017observability}
\begin{equation}\label{eq:battmodel}
  \dot{\mathbf{b}}(t)\hspace*{-.4ex}=\hspace*{-.7ex}\begin{bmatrix}-1/(R_1C_1)\hspace*{-2ex}&\hspace*{-2ex}0&0\\\hspace*{2ex}0&\hspace*{-3ex}-1/(R_2C_2)&0\\\hspace*{2ex}0&\hspace*{-2ex}0&0\end{bmatrix}\hspace*{-.6ex}\mathbf{b}(t)\hspace*{-.2ex}+\hspace*{-.6ex}\begin{bmatrix}\hspace*{-.2ex}1/C_1\\\hspace*{-.2ex}1/C_2\\-\hspace*{-.4ex}\zeta/Q\end{bmatrix}\hspace*{-.7ex}I(t),
\end{equation}
where $\zeta\in\mathbb{R}$ is a battery coefficient~\cite{seewald2022energy}, $R_1,R_2\in\mathbb{R}$ and $C_1,C_2\in\mathbb{R}$ are resistors and capacitors relative to the first and second RC elements in the ECM measured in ohms and farads respectively.
$Q\in\mathbb{R}$ is the battery nominal capacity measured in amperes per hour. 
$I\in\mathbb{R}$ is the internal current which is load-dependent, e.g., the current required to run the motors, actuators, etc. It is measured in amperes and assumed constant in flight from empirical observations of the aerial robot used in the experiments (see Sec.~\ref{sec:res}).

The state $\mathbf{b}:=\begin{bmatrix}V_1&V_2&\mathbf{b}_{\text{SoC}}\end{bmatrix}\in\mathbb{R}^3$ contains three battery metrics. $V_1,V_2\in\mathbb{R}$ are the voltages measured in volts across the first and second RC elements, and $\mathbf{b}_{\text{SoC}}\in(0,1]$ is the normalized battery SoC that evolves from fully charged -- or from a given initial value $\mathbf{b}_{\text{SoC}}(t_0)$ -- to discharged. 
Battery voltage $V_e\in\mathbb{R}$ measured in volts can be 
expressed 
\begin{equation}\label{eq:battvolt}
  V_e(t)=V\big(\mathbf{b}_{\text{SoC}}(t)\big)-V_1(t)-V_2(t)-I(t)R,
\end{equation}
where $R\in\mathbb{R}$ is the single resistor measured in ohms in Fig.~\ref{fig:thevenin}, and $V$ is the open circuit voltage that 
can be retrieved from the 
datasheet~\cite{hinz2019comparison}.

\begin{figure}[t!]
  \begin{minipage}[t!]{.5\columnwidth}
    \vspace*{-.2cm}
    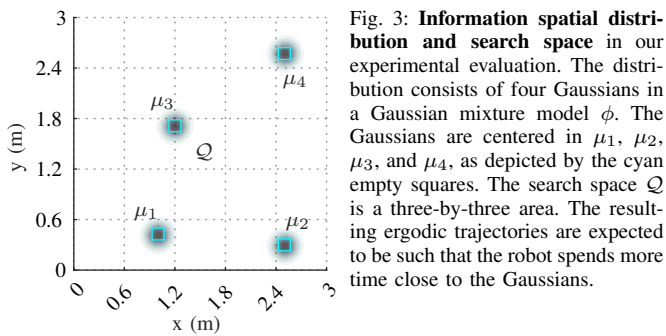
  \end{minipage}
  \begin{minipage}[c]{.48\columnwidth}
    \vspace*{.05cm}
    \caption{\textbf{Information spatial distribution and search space} in our experimental evaluation. The distribution consists of four Gaussians in a Gaussian mixture model $\phi$. The Gaussians are centered in $\mu_1$, $\mu_2$, $\mu_3$, and $\mu_4$, as depicted by the cyan empty squares. The search space $\mathcal{Q}$ is a three-by-three area. The resulting ergodic trajectories are expected to be such that the robot spends more time close to the Gaussians.}
    \label{fig:scenario}
  \end{minipage}
  \vspace*{-.15cm}
\end{figure}

\begin{figure}[b!]
  \vspace*{-.15cm}
  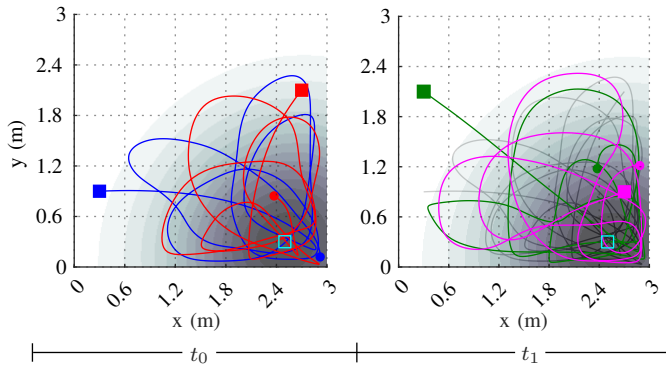
  \caption{\textbf{Experimental evaluation of 
  competing exploration with one Gaussian}.
  Four agents $\alpha_1$, $\alpha_2$, $\alpha_3$, and $\alpha_4$ explore the space two-by-two first, and they compete for one area with high information density. The agents $\alpha_1$ blue and $\alpha_2$ red explore the space in the first horizon $t_0$ (left of the figure), spending most of the time close to the Gaussian. The agents then return to the charging station to recharge the battery. The other two agents $\alpha_3$ dark green and $\alpha_4$ magenta proceed in the next time horizon.}
  \label{fig:res2}
\end{figure}

\begin{figure}[b!]
  \vspace*{-.15cm}
  \hspace*{-.2cm}
  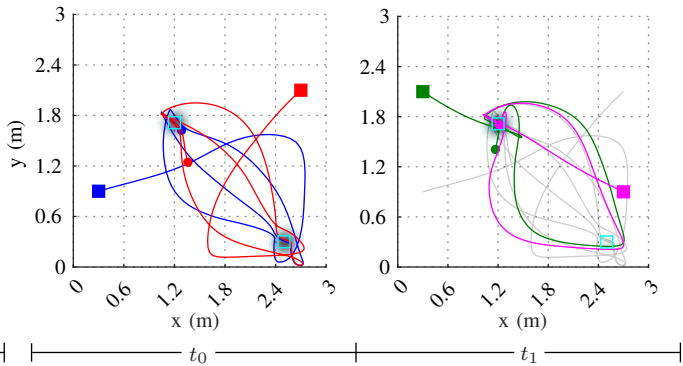
  \caption{\textbf{Experimental evaluation of 
  competing exploration with two Gaussians}. Four agents explore the space and compete for two areas with high information density (instead of one in Fig.~\ref{fig:res2}). The agents blue and red are selected first. One can note how both agents swap between the areas but spend most time near the Gaussians. At the end of the first horizon, they return to the charging stations with the other two agents dark green and magenta exploring the space.}
  \label{fig:res3}
\end{figure}

\begin{figure*}[t!]
  \vspace*{.17cm}
  \begin{minipage}[t]{1\columnwidth}
    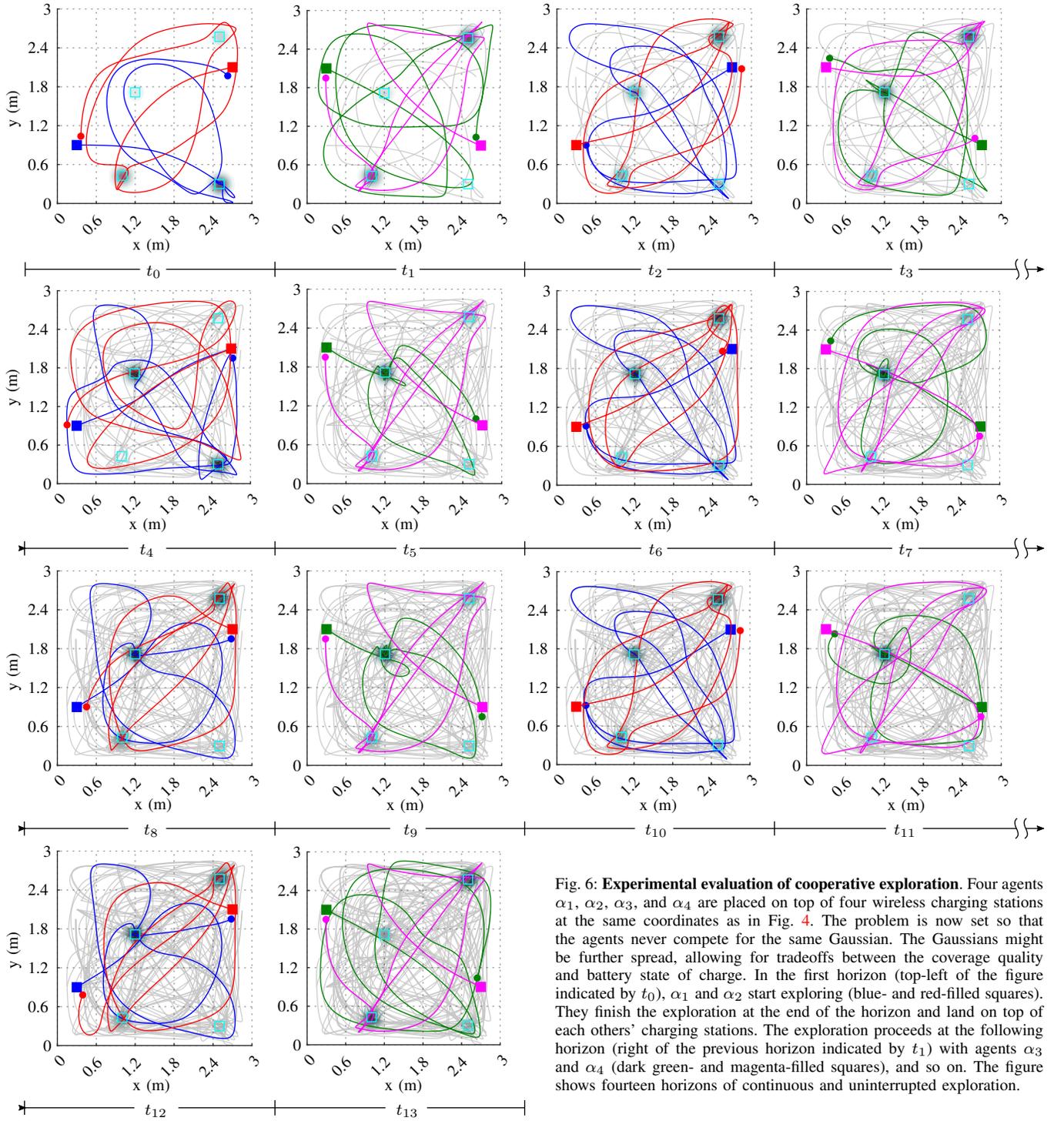
  \end{minipage}
  \hspace{.42cm}
  \begin{minipage}[t]{.97\columnwidth}
    \vspace*{-4.2cm}
    \caption{\textbf{Experimental evaluation of cooperative exploration}. Four agents $\alpha_1$, $\alpha_2$, $\alpha_3$, and $\alpha_4$ are placed on top of four wireless charging stations at the same coordinates as in Fig.~\ref{fig:res2}. The problem is now set so 
    that the agents never compete for the same Gaussian. The Gaussians might be further spread, allowing for tradeoffs between the coverage quality and battery state of charge. In the first horizon (top-left of the figure indicated by $t_0$), $\alpha_1$ and $\alpha_2$ start exploring (blue- and red-filled squares). They finish the exploration at the end of the horizon and land on top of each others' charging stations. The exploration proceeds at the following horizon (right of the previous horizon indicated by $t_1$) with agents $\alpha_3$ and $\alpha_4$ (dark green- and magenta-filled squares), and so on. The figure shows fourteen horizons of continuous and uninterrupted exploration.}
    \label{fig:res}
  \end{minipage}
  \vspace*{-.6cm}
\end{figure*}

The values of $R_1,C_1,R_2,C_2,R$ are identified so that the model output and the physical behavior of the agents are matched as closely as possible~\cite{zhao2017observability} (see Sec.~\ref{sec:res}).

The battery model 
can be used to derive the battery SoC and voltage in Eq.~(\ref{eq:battmodel}) and (\ref{eq:battvolt}). The battery SoC depends on $Q$ and $\zeta$ and is utilized to find the control action $\mathbf{u}(t)$ so that there is at least one agent exploring. 
The voltage depends on the two RC elements and is utilized to evaluate the SoC of a physical system, i.e., it can be compared to the battery voltage provided by a flight controller. 
When the solution of the OCP in Eq.~(\ref{eq:ocpfinal}) is evaluated, the battery model in Eq.~(\ref{eq:battmodel}) is integrated for the duration of the horizon. The recharging is approximated with the expression $\mathbf{b}_{\text{SoC}}=\eta\,\mathbf{b}_{\text{SoC}}+\theta$ for given $\eta,\theta\in\mathbb{R}$ determined empirically.

\section{Experimental Results}\label{sec:res}
\noindent
In this section, we discuss our experimental setup and results. Our experiments are implemented in simulation using  \textsc{Matlab} (R), and physical experiments  are implemented in Python and conducted using Crazyflie 2.0 micro aerial vehicles (MAVs). 
The dynamics $\dot{\mathbf{q}}=f(\mathbf{q},\mathbf{u})$ is that of a 2D single integrator system, which mimics the MAV control reasonably~\cite{dong2023time}. The chosen dynamics is not specific to our implementation. The physical setup is illustrated in Fig.~\ref{fig:abs}.

The source code\cref{reflink} is released under the 
non-commercial open-source license CC BY-NC-SA~4.0. The solution of the OCP in Eq.~(\ref{eq:ocpfinal}) relies on two external open-source components from the literature: the popular nonlinear programming solver IPOPT~\cite{wachter2006implementation} and a software framework for nonlinear optimization called CasADi~\cite{andersson2012casadi}. The simulation is derived offline first, but the computational load is not prohibitive, i.e., online runtime is possible (see Sec.~\ref{sec:conc}). 
Each MAV is equipped with a positioning and wireless charging decks. Precise positioning of MAVs is achieved via two HTC SteamVR Base Station 2.0 units. Each MAV is then equipped with a one-cell 250 mAh 3.7 volts LiPo battery.

We evaluate our approach under two different scenarios. In both scenarios, we use a three-by-three-meter space where the knowledge of the environment is assumed. The spatial distribution $\phi$ contains four Gaussians in the GMM in Eq.~(\ref{eq:gmm}) in the second scenario, as illustrated in Fig.~\ref{fig:scenario} (the four cyan empty squares). In the first scenario, it contains one Gaussian centered in $\mu_2$ first, and it contains two Gaussians centered in $\mu_2$ and $\mu_3$ later.

\subsection*{Competing exploration}
\noindent
In the first scenario, four MAVs $\alpha_1$, $\alpha_2$, $\alpha_3$, and $\alpha_4$ are placed on top of four wireless charging stations. 
The horizon is set to two and a half minutes and is derived empirically along with battery and recharging coefficients. The battery values used in the scenario are scaled from~\cite{zhao2017observability}. The number of frequencies $K$ is set to nine, as in~\cite{calinon2020mixture}. 
The MAVs compete for the same area with high information density, i.e., they utilize the multi-agent ergodic metric introduced in Eq.~(\ref{eq:multiergometric}). 
The battery constraint is edited so that there are two MAVs first, i.e.,\vspace*{-.165cm}
\begin{equation}
\exists_{=1}\, k_1,k_2\in[n]\text{ s.t. }{}^{k_1} \mathbf{b}_{\text{SoC}},{}^{k_2} \mathbf{b}_{\text{SoC}}\in(0,b_f],\vspace*{-.1cm}
\end{equation} 
where 
$\exists_{=1}$ indicates the unique existential quantification.

Initially, two MAVs are selected via the solution to the OCP in Eq.~(\ref{eq:ocpfinal}), $\alpha_1$ ``blue'' and $\alpha_2$ ``red.'' They are located at coordinates $($0.3$,$0.9$)$ and $($2.7$,$2.1$)$ respectively, denoted by the blue and red filled squares in Fig.~\ref{fig:res2}. The MAVs explore the space for the first horizon, focusing on the area with high information density. At the end of the horizon (blue and red filled dots), the MAVs return to the charging stations. 

Once the two agents $\alpha_1$ and $\alpha_2$ land, they start recharging. The formulation of the OCP in Eq.~(\ref{eq:ocpfinal}) is such that the other two agents $\alpha_3$ ``dark green'' and $\alpha_4$ ``magenta'' are selected. They are located at coordinates $($0.3$,$2.1$)$ and $($2.7$,$0.9$)$. 
In the following horizon, agents $\alpha_3$ and $\alpha_4$ are recharging whereas $\alpha_1$ and $\alpha_2$ proceed with the exploration, 
and so~on.

Fig.~\ref{fig:res3} illustrates the case of the agents competing for two areas with high information density.

\begin{figure}[t!]
  \begin{minipage}[t!]{.5\columnwidth}
    \vspace*{-.2cm}
    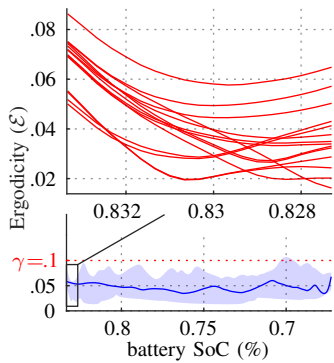
  \end{minipage}
  \begin{minipage}[c]{.48\columnwidth}
    \vspace*{.05cm}
    \caption{\textbf{Ergodicity 
    as a function of state-of-charge}. The top plot shows the evolution of the ergodicity for all the horizons in Fig.~\ref{fig:res}. The bottom shows the average ergodicity. Initially, the MAVs are at their charging stations. As they start exploring, they move to the areas with high information density -- ergodicity decreases. As they approach the end of the horizon, they start moving to the charging stations -- ergodicity increases. The average ergodicity can be observed to be under the value $\gamma$.}
    \label{fig:ergo}
  \end{minipage}
  \vspace*{-.4cm}
\end{figure}

\subsection*{Cooperative exploration}
\noindent
In the second extensive scenario, four MAVs $\alpha_1$, $\alpha_2$, $\alpha_3$, and $\alpha_4$ are placed on top of four wireless charging stations, as in the previous scenario. The optional constraints in Eq.~(\ref{eq:ocpfinalextra1}) are built so that each MAV covers two Gaussians at a time that are respectively farthest ($\mu_3$ and $\mu_2$ are the centers of the Gaussians covered by the blue and dark green agents, $\mu_1$ and $\mu_4$ are the centers of the Gaussians covered by the red and magenta agents in Fig.~\ref{fig:res}). This means that the MAVs will never compete for the same Gaussian, but will cooperate in the exploration. 

There is an additional constraint on the final point in Eq~(\ref{eq:ocpgivenconst}), set so that the agents have to be in the proximity of a charging station. The actual constraint is derived in Eq.~(\ref{eq:resaddconst}). 
The cost function in Eq.~(\ref{eq:ocpfinalcost}) is further enhanced with the mixing coefficient $\boldsymbol{\delta}$ in Eq.~(\ref{eq:gmm}), allowing us to find the tradeoffs between the single Gaussians, the different agents, and the battery SoC. Namely, the cost is
\begin{equation}
  \min_{\boldsymbol{\Theta},\boldsymbol{\delta}}\,\,\,{\frac{1}{n}\sum_{k=1}^{n}\int_{\mathcal{T}_k}{}^k\hspace*{-.1ex}\mathbf{u}(\tau)^TR_k\,{}^k\hspace*{-.1ex}\mathbf{u}(\tau)\,d\tau-\sum_{k=1}^{m}\delta_k}.
\end{equation}

A similar approach is undertaken in prior literature~\cite{rao2023multi}, where the ergodic objective is dynamic as more information is gathered, rather than the battery status is changed.

The number of frequencies, battery and recharging coefficients, and the horizon are those used in the previous scenario.
The ergodic metric 
is set to be lower or equal to 0.1, 
in line with similar literature~\cite{dong2023time}.

The results are shown in Fig.~\ref{fig:res}. The figure is to be read from left to right and from top to bottom, with the horizons being indicated under each subfigure (meaning that $t_0$ is the first horizon, $t_1$ is the second horizon, etc.). Initially, two MAVs are selected via the solution to the OCP in Eq.~(\ref{eq:ocpfinal}), $\alpha_1$ blue and $\alpha_2$ red, similarly to the previous scenario. 
The energy-aware ergodic trajectories ${}^1\mathbf{q}(t)$ and ${}^2\mathbf{q}(t)$ are selected so that the MAVs land at each other's charging stations, i.e., ${}^1\mathbf{q}(t_f)={}^2\mathbf{q}(t_0)$ and vice-versa. The mixing coefficients for $\alpha_1$ are such that $\delta_2>\delta_3$, meaning that the agent $\alpha_1$ explores in more detail the area delimited by the Gaussian centered in $\mu_2$. This is indicated by the darker coloring of the different Gaussians, which is proportional to the optimal value of $\boldsymbol{\delta}$. 
An analogous situation is to be observed with agent $\alpha_2$. 

To guarantee that both agents land 
on top of each other's charging stations (the red and blue filled dots at the end of the trajectories for $\alpha_2$ and $\alpha_1$) respectively, 
the constraint in Eq.~(\ref{eq:ocpgivenconst}) is evaluated within
\begin{equation}\label{eq:resaddconst}
  \lVert{}^{k_2}\mathbf{q}(t_f)-{}^{k_1}\mathbf{q}(t_0)\rVert\leq\varepsilon,
\end{equation} 
where $\varepsilon\in\mathbb{R}_{>0}$ and $k_1,k_2\in[n]$ are given.

Once the two agents $\alpha_1$ and $\alpha_2$ land, they start recharging. 
The other two agents $\alpha_3$ dark green and $\alpha_4$ magenta are selected. 
They proceed on the respective energy-aware ergodic trajectories and land on top of each other's charging stations, with the past trajectory being indicated in the background in gray. The figure shows fourteen horizons. 

\subsection*{Ergodicity against battery state of charge}
\noindent
We report the evolution of the value of the ergodic metric in Eq. (\ref{eq:ergmetric}) in time as a function of the battery SoC in Fig.~\ref{fig:ergo}. The experimental data are from the cooperative exploration.

The top of the figure shows the evolution per each horizon in red, whereas the bottom shows the average value in blue. We can observe that the exploration starts at an initial value of ergodicity, which mostly depends on the distance from the charging stations to the components of the GMM (i.e., high information density). The ergodicity decreases as the agents move towards the Gaussians in the spatial distribution GMM. It oscillates as the agents move from one Gaussian to another. In the first half of the horizon, the average ergodicity continues to descend as more information is gathered. In the second, the ergodicity increases, peaking at the end, as the discharged agents return to the charging stations for recharging (i.e., low information density).

\section{Conclusion and Future Directions}\label{sec:conc}
\noindent
This work enhances prior literature on ergodic search and 
answers the question of whether is it possible to explore a space uninterruptedly, with at least one agent at all times. 
Our methods are to derive an abstract battery model and extend the canonical ergodic search -- a method to derive robots' trajectories that visit areas with high information density -- to energy-aware ergodic search. Continuous exploration is achieved using an 
optimization framework, which resembles a model predictive controller formulation. 
Experimental data indicate the effectiveness of our battery-constrained exploration. Continuous and uninterrupted coverage is achieved with a multi-agent system so that there is always at least one agent exploring and the spatial distribution is satisfied -- a statement that we prove with empirical evidence.

A limitation of the current methods is that the charging stations are in fixed positions and the information is centralized. To enable real-world use cases, we are currently extending the methods to mobile charging stations and decentralized 
systems, which arise in scenarios such as environmental surveying. In future work, we are also planning to investigate other aspects, including energy optimality, online runtime, etc., which are currently not addressed. 

{\small
\bibliographystyle{IEEEtran} 
\bibliography{enerergo}
}

\end{document}